\documentclass{article}
\usepackage{spconf,amsmath,graphicx}
\usepackage{spconf,amsmath,graphicx}
\usepackage{acronym} 
\usepackage{booktabs}
\usepackage[export]{adjustbox}
\usepackage{multirow}
\usepackage{subcaption}
\usepackage[font=small,skip=0pt]{caption}
\newacro{fgcsei}[FGC-SEI]{Film Grain Characteristics Supplemental Enhancement Information}
\newacro{inn}[INN]{Invertible Neural Network}
\newacro{fgann}[FGA-NN]{Film Grain Analysis Neural Network}
\newacro{mse}[MSE]{Mean Square Error}
\newacro{mae}[MAE]{Mean Absolute Error}
\newacro{vvc}[VVC]{Versatile Video Coding}
\newacro{vtm}[VTM]{VVC Test Model}
\newacro{hevc}[HEVC]{High Efficiency Video Coding}
\newacro{jsdnss}[JSD-NSS]{Jensen Shannon divergence - natural scene statistics}
\newacro{lpips}[LPIPS]{learned perceptual image patch similarity}
\newacro{kld}[KLD]{KL divergence}
\newacro{vfgs}[VFGS]{Versatile Film Grain Synthesis}
\newacro{gan}[GAN]{Generative Adversarial Network}

\usepackage{xcolor}

\title{FGA-NN: Film Grain Analysis Neural Network}

\name{Zoubida AMEUR, Frédéric LEFEBVRE, Philippe DE LAGRANGE and Miloš RADOSAVLJEVIĆ}
\address{InterDigital R\&D,  Cesson-S\'evign\'e, France\\
firstname.lastname@interdigital.com}
 
\begin{document}

\maketitle

\begin{abstract}
Film grain, once a by-product of analog film, is now present in most cinematographic content for aesthetic reasons. However, when such content is compressed at medium to low bitrates, film grain is lost due to its random nature. To preserve artistic intent while compressing efficiently, film grain is analyzed and modeled before encoding and synthesized after decoding. This paper introduces FGA-NN, the first learning-based film grain analysis method to estimate conventional film grain parameters compatible with conventional synthesis. Quantitative and qualitative results demonstrate FGA-NN's superior balance between analysis accuracy and synthesis complexity, along with its robustness and applicability.

\end{abstract}

\begin{keywords}
film grain, neural network, video compression.
\end{keywords}
\section{Introduction}
\label{sec:intro}
Film grain, appreciated in video production for its natural look and creative expression, originates from the physical exposure and development of photographic film. Unlike film, digital sensors do not undergo such a process and are therefore grain-free. To add texture, warmth or evoke nostalgia, filmmakers often reintroduce grain into digital content in a content-specific manner adapted to factors such as pixel intensity. The random nature of film grain, however, poses a major challenge to conventional video codecs, which often eliminate grain at medium to low bitrates, thus compromising the visual quality and the artistic intent of the content. Conversely, preserving film grain requires disproportionate bitrates which results in an inefficient compression. 

To efficiently preserve film grain, in state-of-the-art video codecs like \ac{vvc} \cite{vvc}, an alternative to high bitrate encoding is proposed. It consists in analyzing and estimating film grain parameters prior to encoding and synthesizing it back after decoding using the estimated parameters transmitted as metadata. \ac{vvc} natively supports the signaling of film grain parameters as metadata through a well-defined \ac{fgcsei} message \cite{sei, milos}. This paper primarily addresses the film grain analysis stage.

In conventional codecs, the film grain analysis workflow follows a standard process. First, denoising is applied to extract the film grain image by finding the difference between the grainy source and its denoised version. Further analysis focuses on features such as edges and texture, limiting the analysis to flat, uniform regions. Based on this pre-processing, model parameters are determined, including \textit{grain amplitude} and \textit{grain pattern}. 
Grain amplitude is determined in relation to image intensity using some scaling function while grain pattern is modeled by identifying frequency limits (cut-off frequencies) for frequency-based models or auto-regressive parameters. Consequently, film grain synthesis is typically based on the generation of Gaussian noise, with spatial correlation modeled by frequency limits or auto-regressive parameters, and local adaptation consisting of adjusting grain amplitude. Conventional methods of film grain analysis show some limitations, as the accuracy of the estimation of film grain parameters is highly dependent on the result of denoising and edge detection processing. Moreover, considering only homogeneous blocks in the analysis step, limits the data available for processing, consequently leading to reduced synthesis accuracy.

Alongside conventional approaches to film grain analysis and synthesis, learning-based frameworks have emerged. Style-FG \cite{stylefg} is the first deep learning framework for film grain analysis and synthesis with grain characteristics encoded as a latent vector representation. 3R-INN \cite{3rinn} is another learning-based framework that is based on \acp{inn}. Thanks to its invertibility, 3R-INN performs analysis in a forward pass where grain information is captured in a latent variable constrained to follow a standard Gaussian distribution and performs synthesis in an inverse pass without requiring explicit metadata. Learning-based methods offer higher accuracy but produce parameters incompatible with video coding standards, in addition to necessitating complex synthesis modules. This incompatibility makes their integration into practical video coding systems challenging, particularly for resource-constrained devices.

\begin{figure*}[t!]
    \centering
    \includegraphics[width=.8\textwidth]{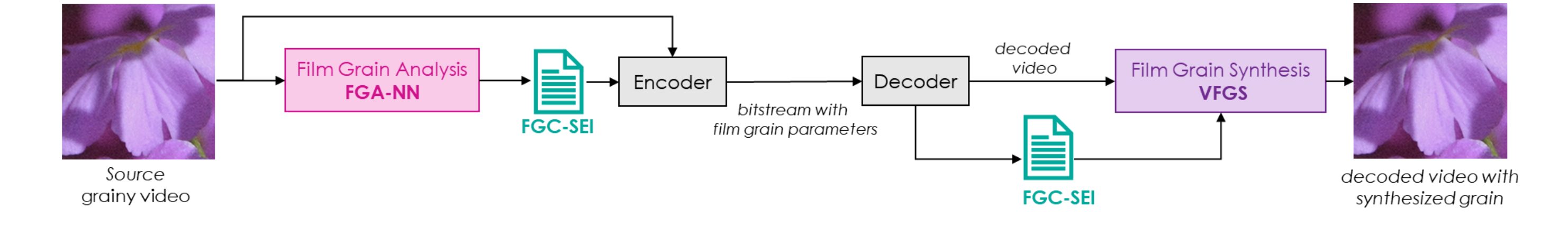}
    \captionsetup{aboveskip=-0pt, belowskip=-15pt}    
    \caption{Framework for proposed film grain analysis and synthesis workflow in a video distribution system, utilizing FGA-NN for analysis combined with VFGS for synthesis.}
    \label{fig:framework}
\end{figure*}

By leveraging the accuracy of learning-based models and the compatibility of conventional methods, this paper introduces \ac{fgann} which analyzes film grain from grainy videos and provides film grain characteristics in the format supported by recent video codecs, known as FGC-SEI. FGA-NN along with the new \ac{fgcsei} dataset are detailed in Section \ref{sec:method}. Section \ref{sec:experimental} presents a comparative of the performance of FGA-NN against state-of-the-art methods. Finally, conclusions and perspectives are discussed in Section \ref{sec:conclusion}.

\section{Proposed method}
\label{sec:method}
\subsection{System overview}
Given a grainy video as input, FGA-NN analyzes the film grain and outputs film grain parameters in the FGC-SEI format supported by recent video coding standards. The video is then encoded and transmitted with these parameters. On the decoder side, the video is decoded, and film grain is synthesized using \ac{vfgs} \cite{vfgs}, the conventional film grain synthesis method, utilizing the transmitted FGC-SEI parameters, as illustrated in Figure \ref{fig:framework}. To train FGA-NN, a dataset of grainy videos (inputs) paired with their corresponding FGC-SEI parameters (outputs) is required. Typically, only grainy videos are available, and for synthetic grain, the parameters of the post-production tool might be known but not in FGC-SEI format. FGC-SEI parameters can be derived either manually which may be difficult and time consuming or by using a film grain analysis module designed for video coding, underscoring the need for FGA-NN and such a dataset. The following subsection details the process of the dataset creation.

\subsection{FGC-SEI Dataset}
\label{sec:dataset}
To create the necessary pairs (grainy videos, FGC-SEI parameters), we approached the problem inversely by first generating a set of FGC-SEI parameters which we then used to add film grain to clean (grain-free) videos using \ac{vfgs}\cite{vfgs}. Table \ref{table:fgc_sei_parameters} reports key parameters used to create our dataset. 
The FGC-SEI message supports two synthesis approaches (\textbf{SEIFGCModelID}): frequency filtering modeling (0) and auto-regressive modeling (1). Currently, only the frequency model is implemented in codec software like \ac{vtm} reference software \cite{VTMsoft} or the real-world Versatile Video Encoder and Decoder implementation software \cite{VVenC,VVdeC}. Therefore, FGA-NN aligns with and provides parameters for the frequency filtering model. As for blending mode (\textbf{SEIFGCBlendingModeID}), VFGS uses additive blending mode (0). In the frequency filtering model, each color component (Y:0, Cb:1, Cr:2) for which synthesis process is invoked, a number of intensity intervals is defined \textbf{SEIFGCNumIntensityIntervalMinus1Comp0,1,2} wherein lower and upper bounds are defined in \textbf{SEIFGCIntensityIntervalLowerBoundComp0,1,2} and \textbf{SEIFGCIntensityIntervalUpperBoundComp0,1,2}, respectively. The number of model parameters \textbf{SEIFGCNumModelValuesMinus1Comp0,1,2} is limited to 3, limiting it to scale [0-255] and high vertical and horizontal frequency cut-offs [2-14]. \textbf{SEIFGCLog2ScaleFactor} indicates the scale of the scaling factors. Acting on this parameter is a quick way of changing the film grain strength.

\begin{table}[h!]
\centering
\caption{FGC-SEI Parameters}
\begin{adjustbox}{max width=.6\linewidth}
\begin{tabular}{|l|c|}
\hline
\textbf{Parameter} & \textbf{Value} \\ \hline
\textbf{SEIFGCModelID} &0 \\ \hline
\textbf{SEIFGCBlendingModeID}&0 \\ \hline
\textbf{SEIFGCLog2ScaleFactor}&2-7    \\ \hline
\textbf{SEIFGCCompModelPresentComp0,1,2} &1  \\ \hline
\textbf{SEIFGCNumIntensityIntervalMinus1Comp0,1,2} &0-255\\ \hline
\textbf{SEIFGCIntensityIntervalLowerBoundComp0,1,2} &0-255 \\ \hline
\textbf{SEIFGCIntensityIntervalUpperBoundComp0,1,2} &0-255 \\ \hline
\textbf{SEIFGCNumModelValuesMinus1Comp0,1,2}&2 \\ \hline
\end{tabular}
\label{table:fgc_sei_parameters}
\end{adjustbox}
\end{table}

\begin{figure*}[t!]
    \centering
    \begin{subfigure}[]{0.3\linewidth}
            \includegraphics[width=\linewidth]{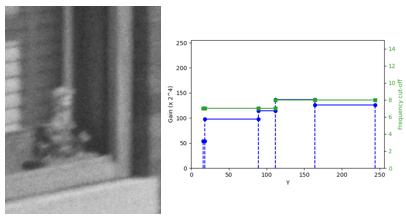}
            \caption*{Ground-truth}
    \end{subfigure}     
    \begin{subfigure}[]{0.3\linewidth}
            \includegraphics[width=\linewidth]{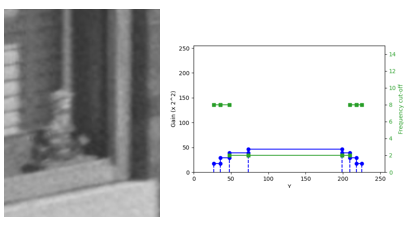}
            \caption*{FGA-CONVENT}
    \end{subfigure}    
    \begin{subfigure}[]{0.3\linewidth}
            \includegraphics[width=\linewidth]{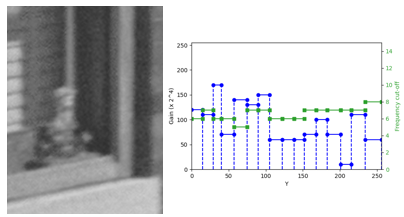}
            \caption*{FGA-NN}
    \end{subfigure}    
    \captionsetup{aboveskip=-0pt, belowskip=-15pt}    
    \caption{Comparison of grainy images and their Luma FGC-SEI parameters (AmericanChoice sequence).}
    \label{fig:analysis}
\end{figure*}
 
To build a large and diverse dataset, we randomly generated 300 set of FGC-SEI parameters while adhering to specific constraints to ensure visually accurate film grain. We used three different Log2scale factors [3-5] and defined 16 intervals for luma and 6 for chroma. For each interval, we selected a scaling factor (a multiple of 10 within the range of 0 to 255) and a single cut-off frequency (the same for both vertical and horizontal) within the range of 3 to 14 for luma and 4 to 8 for chroma. Also, we imposed specific constrains to avoid drastic changes in grain amplitude (scale factor) and grain pattern (cut-off frequencies) between consecutive intervals.

These FGC-SEI parameters were used to add grain on a set of clean videos grouping the BVI-DVC \cite{bvi} and the DIV2K \cite{div2k} datasets.
\subsection{Network architecture}
Given a grainy video as input, two versions of the FGA-NN module processes separately luma and chroma channels, predicting each, intervals boundaries (lower and upper), their associated scaling factors, cut-off frequencies, and a global Log2Scale factor. The models address these prediction tasks as one regression and three classification problems. Interval boundary prediction is formulated as a regression task, with outputs ranging from 0 to 255. Cutoff frequencies, scaling factors, and Log2Scale factor predictions are treated as classification tasks.

\begin{figure}[h!]
    \centering
    \includegraphics[width=\linewidth]{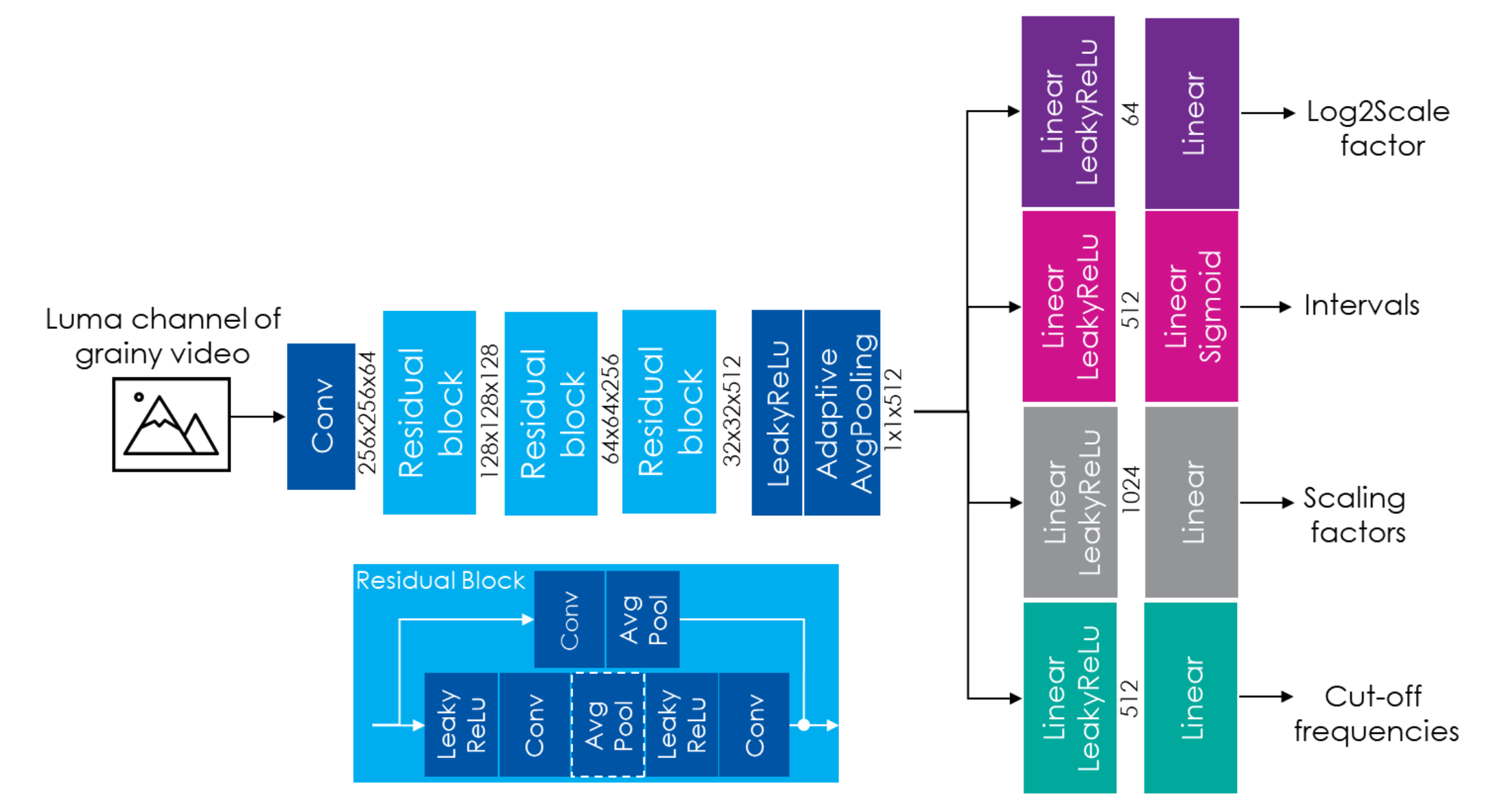}
    \captionsetup{aboveskip=-0pt, belowskip=-20pt}    
    \caption{Luma FGA-NN detailed architecture.}
    \label{fig:nn}
\end{figure}

Both luma and chroma versions of FGA-NN employ a shared backbone to extract features from grainy image inputs followed by four parallel task-specific heads to map these features to the desired outputs. The backbone of FGA-NN comprises one convolutional layer followed by three residual blocks and an adaptive average pooling while task-specific heads comprise each two linear layers. Linear layers dimensions are adapted to prediction task complexity.  Specifically, to predict Log2Scale factor (3 possible classes: 2-5) 64-dimensional layers are used, to predict scaling factors (26 possible classes: multiples of 10 in 0-255) 1024-dimensional layers are used, and to predict cut-off frequencies (12 possible classes for luma: 3-14, and 5 for chroma: 3-7) 512-dimensional layers are used. Fig. \ref{fig:nn} illustrates the luma FGA-NN detailed architecture. Note that the chroma version of FGA-NN is tailored to adapt the input and output dimensions accordingly. 

\subsection{Training details}
FGA-NN is trained using a combination of four training objectives, each corresponding to one of the four tasks. The classification tasks ($*$) (cut-off frequencies, scaling factors, and Log2Scale factor predictions) are trained by minimizing the categorical cross-entropy loss, which measures the discrepancy between the predicted and actual labels and is formulated as:
\begin{equation*}
\small
\mathcal{L}_{*}= -\sum_{i=1}^{N} y_i \log(\hat{y}_i)
\end{equation*}
where $y_i$ is the true label, $\hat{y}_i$ is the predicted probability of the true label and $N$ is the batch size. 

Interval boundaries, normalized to [0, 1], are predicted by minimizing a combined loss function: \textbf{an exponentially-scaled L1 loss} $exp_{L1}$ which heavily penalizes large prediction errors and \textbf{a monotonicity penalty loss} $mono_{P}$ which enforces non-decreasing behavior by calculating differences between consecutive elements and penalizing only the negative ones, as follows:
\begin{gather*}
\small
exp_{L1}= = \frac{1}{N} \sum_{i=1}^{N} \left( e^{\beta \cdot |y_i - \hat{y}_i|} - 1 \right)  \\ \nonumber
mono_{P} = \frac{1}{N} \sum_{j=1}^{N} \max(-(\hat{y}_{i+1} - \hat{y}_i), 0) \\ \nonumber
\mathcal{L}_{Intervals} = exp_{L1} + mono_{P} 
\end{gather*}
with $\beta$ being a hyper-parameter controlling the sensitivity to the error and set to 5.

The total loss function combines the individual losses, weighted by their respective regularization parameters: 

\begin{equation}
\small
\mathcal{L}=  \lambda_1 \mathcal{L}_{Cut-off}+\lambda_2 \mathcal{L}_{Intervals} + \lambda_3 \mathcal{L}_{Log2Scale} + \lambda_4 \mathcal{L}_{scaling}
\end{equation}
 with $\lambda_1$ and $\lambda_4$ to 100 and $\lambda_2$ set to 1 and $\lambda_3$ set to 0.1.\\

\begin{figure*}[t!]
    \centering
    \begin{subfigure}[]{0.15\linewidth}
            \includegraphics[width=\linewidth]{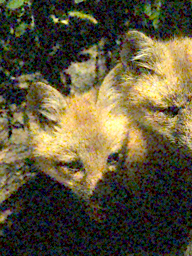} 
            \includegraphics[width=\linewidth]{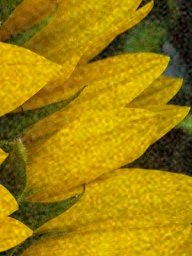}             
            \caption*{Ground-truth}
    \end{subfigure}     
    \begin{subfigure}[]{0.15\linewidth}
            \includegraphics[width=\linewidth]{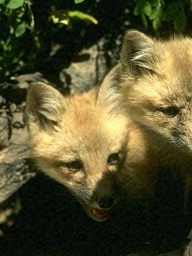} 
            \includegraphics[width=\linewidth]{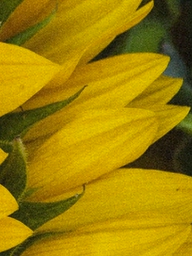}             
            \caption*{FGA-CONVENT}
    \end{subfigure}    
    \begin{subfigure}[]{0.15\linewidth}
            \includegraphics[width=\linewidth]{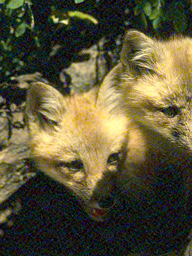} 
            \includegraphics[width=\linewidth]{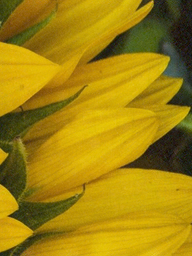} 
            \caption*{Style-FG}
    \end{subfigure} 
    \begin{subfigure}[]{0.15\linewidth}
            \includegraphics[width=\linewidth]{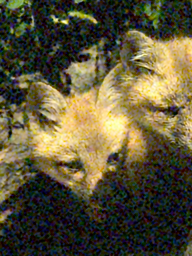} 
            \includegraphics[width=\linewidth]{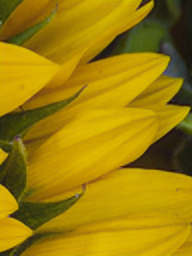}             
            \caption*{3R-INN}
    \end{subfigure} 
    \begin{subfigure}[]{0.15\linewidth}
            \includegraphics[width=\linewidth]{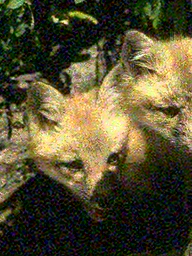} 
            \includegraphics[width=\linewidth]{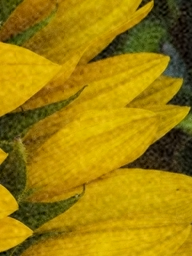}             
            \caption*{FGA-NN}
    \end{subfigure} 
    \captionsetup{aboveskip=-0pt, belowskip=-15pt}            
    \caption{Comparison of synthesized grainy images using different analysis and synthesis workflows to ground-truth ones.}
    \label{fig:qualitative_comparaison}
\end{figure*}

Adam optimizer \cite{Adam1, Adam2} with $ \beta_{1} = 0.9$ and $\beta_{2} = 0.999$ is used. Weights are initialized using the classic He initialization \cite{he2015delving}. The batch size is set to $64$ and  the learning rate is set to $5\mathrm{e}{-4}$, with a total of $10$k training iterations.

\section{Experimental results}
\label{sec:experimental}
FGA-NN aims to accurately estimate film grain parameters from grainy videos for efficient compression and faithful synthesis. In the following subsections, we evaluate FGA-NN performance in terms of accuracy of estimated parameters, synthesis fidelity and bitrate savings.

\subsection{Film grain analysis evaluation}
\label{subsec:tool}
This subsection presents an evaluation of FGA-NN in comparison to FGA-CONVENT (\cite{VTMsoft}), the only existing state-of-the-art approach capable of estimating parameters compatible with the FGC-SEI format. UHD sequences with real film grain from the JVET subjective evaluation test set \cite{jvet} are used for this evaluation. Fig. \ref{fig:analysis} compares the same cropped region from the grainy synthesized videos generated by VFGS using parameters estimated by FGA-CONVENT and FGA-NN, against the ground truth. It also contrasts their estimated luma channel film grain parameters with expert-tuned ground-truth values.
\begin{table}[h!]
\centering
\caption{Quantitative comparison between our method and state-of-the art methods on different test-set.}
\begin{adjustbox}{max width=\linewidth}
\begin{tabular}{l|ccc|ccc}
\toprule
\multirow{2}{*}{Method} & \multicolumn{3}{c|}{FGA-SEI test-set} & \multicolumn{3}{c}{FilmGrainStyle740k test-set\cite{3rinn}}  \\ 
                        & LPIPS$\downarrow$        & JSD-NSS$\downarrow$  &KLD $\downarrow$      & LPIPS$\downarrow$                 & JSD-NSS $\downarrow$  &  KLD $\downarrow$                \\ \midrule
FGA-CONVENT + VFGS \cite{vtm}                   &0.239   &0.0165   &0.039   &0.129   &0.005   &0.016         \\
Style-FG \cite{stylefg}                   &0.228   &0.0156   &0.046   &0.074   &\textbf{0.003}   &\textbf{0.004}    \\
3R-INN \cite{3rinn}                   &0.372   &0.0769   &0.168   &\textbf{0.058}  &0.005   &\textbf{0.004}               \\
FGA-NN + VFGS                   &\textbf{0.073}   &\textbf{0.0023}   &\textbf{0.010}   &0.120   &0.004   &0.010      \\ \bottomrule
\end{tabular}
\label{tab:comparison}
\end{adjustbox}
\end{table}
Film grain parameters are illustrated using an interactive graphical tool \cite{vfgs}, where the X axis represents the pixel values range [0,255]. The blue Y axis represents the scaling factors [0,255] and the green Y axis represents the cut-off frequencies [2,14]. The dashed lines separate the different intervals. As for the Log2Scale factor, it is described in the Y-axis label as Gain($x^{ \textbf{Log2scale}}$). 

One can observe that FGA-NN accurately captures the overall trend of the ground-truth film grain pattern and amplitude, resulting in synthesized images with perceptually similar film grain to that of the ground-truth ones. On the other hand, FGA-CONVENT predicts a lower scaling factor, compensated by a correspondingly lower Log2Scale factor as a result of its design, and tends to generate a coarser film grain pattern than the reference, resulting in a distinct yet visually consistent appearance.
\\ Direct comparison of estimated and ground-truth film grain parameters is challenging. The interplay between scaling factors and Log2Scale factors allows for error compensation. Furthermore, minor variations in the estimated scaling factor or cutoff frequencies are likely to have minimal visual impact.

Furthermore, FGA-NN is trained to provide film grain parameters on the full intensity range [0, 255], while some test data with limited intensity variation may provide insufficient data for accurate prediction. Therefore, the following subsection evaluates the end-to-end analysis and synthesis workflow. Accurate analysis enables faithful synthesis, and comparing grainy images (both subjectively and objectively) is more straightforward.

\begin{figure*}[t!]
    \centering
    \begin{subfigure}[]{0.13\linewidth} 
            \includegraphics[width=\linewidth]{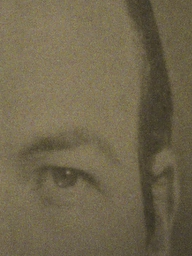}   
            \includegraphics[width=\linewidth]{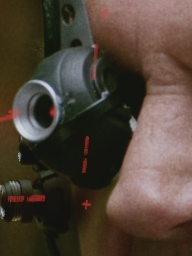}           
            \caption*{Original}
    \end{subfigure} 
    \begin{subfigure}[]{0.13\linewidth} 
            \includegraphics[width=\linewidth]{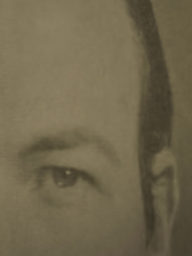}             
            \includegraphics[width=\linewidth]{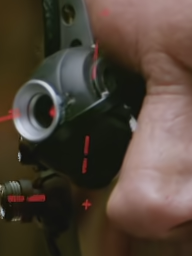} 
            \caption*{Compressed}
    \end{subfigure}         
    \begin{subfigure}[]{0.13\linewidth}
            \includegraphics[width=\linewidth]{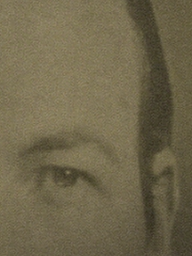}    
            \includegraphics[width=\linewidth]{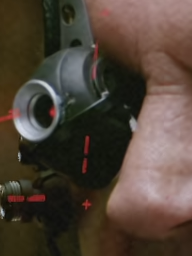}        
            \caption*{FGA-CONVENT}
    \end{subfigure}    
    \begin{subfigure}[]{0.13\linewidth}
            \includegraphics[width=\linewidth]{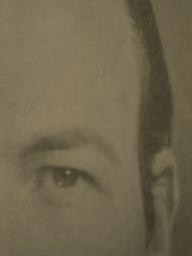} 
            \includegraphics[width=\linewidth]{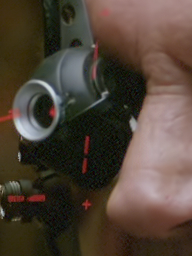} 
            \caption*{Style-FG}
    \end{subfigure} 
    \begin{subfigure}[]{0.13\linewidth} 
            \includegraphics[width=\linewidth]{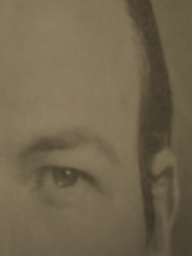}
            \includegraphics[width=\linewidth]{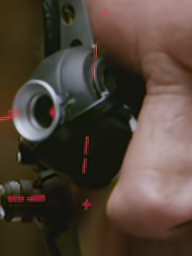} 
            \caption*{3R-INN}
    \end{subfigure} 
    \begin{subfigure}[]{0.13\linewidth}
            \includegraphics[width=\linewidth]{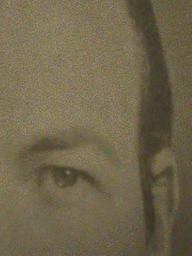}  
            \includegraphics[width=\linewidth]{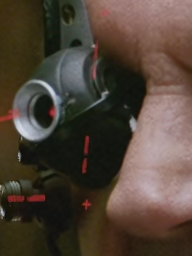}
            \caption*{FGA-NN}
    \end{subfigure} 
    \captionsetup{aboveskip=-0pt, belowskip=-15pt}    
    \caption{Comparison of enhanced compressed images using different analysis and synthesis workflows (third to last column).}
    \label{fig:bitrate}
\end{figure*}

\subsection{Film grain synthesis evaluation}
In this subsection, we evaluate film grain synthesis fidelity by comparing results obtained using: 1) the combination of FGA-NN for analysis and VFGS for synthesis , 2) the combination of FGA-CONVENT for analysis and VFGS for synthesis, 3) Style-FG \cite{stylefg} and 4) 3R-INN \cite{3rinn}. For a fair comparison, methods are evaluated on both the test-set from the FGC-SEI dataset and the test-set from the FilmGrainStyle740k dataset \cite{stylefg}.
Table \ref{tab:comparison} presents a quantitative evaluation of synthesized grainy images against ground-truth images using the \ac{lpips} \cite{lpips}, \ac{jsdnss} \cite{jsdnss}, and \ac{kld} metrics, widely used to evaluate film grain similarity. On the FGC-SEI test-set, synthesis using estimated parameters output by FGA-NN demonstrates superior performance across all metrics. On the FilmGrainStyle740k test-set, Style-FG and 3R-INN achieve the best results, as these methods were specifically trained on this dataset, with FGA-NN trailing closely behind. The performance of FGA-CONVENT combined with VFGS is suboptimal on both test-sets. This is solely due to the fact that the analysis relies on homogeneous regions and exploits information from multiple frames in a real film grain analysis use-case, whereas in the present evaluation analysis is provided with a single low-resolution image (256x256 to a maximum of 768x512), which often contains significant texture. This further complicates the challenge for conventional analysis method, making it impossible to apply FGA-CONVENT to such small images.

Fig. \ref{fig:qualitative_comparaison} presents a subjective visual comparison of synthesized grainy images for a sample from the FilmGrainStyle740k test set (top row) and for a sample from the FGC-SEI test set (bottom row). For the top row sample, synthesis with Style-Fg and 3R-INN is excellent, followed by synthesis using FGA-NN parameters. FGA-CONVENT exhibits the lowest performance as reported by the objective metrics, due to already mentioned limitations. For the bottom row sample, Style-FG and 3R-INN exhibit a significant drop in quality, indicating a sort of overfitting to the FilmGrainStyle740k training set and poor generalization. Synthesis with FGA-NN estimated parameters maintains comparable performance, highlighting its robustness and generalization.

\subsection{Real-world film grain workflow evaluation}
This subsection evaluates different film grain analysis and synthesis workflows on two UHD video sequences (MeridianSmoker1 and TearsOfSteel-044) from the JVET subjective evaluation test set \cite{jvet}, assessing learning-based film grain analysis and synthesis against conventional methods on unseen, real-world content.
Fig. \ref{fig:bitrate} shows cropped regions from different versions of the original video sequences: original (film grain present), compressed at low-bitrate (efficient transmission, film grain lost), and low-bitrate enhanced by the different film grain analysis and synthesis workflows. In the top row sample, both FGA-CONVENT and FGA-NN, when coupled with VFGS, faithfully synthesize film grain, outperforming Style-FG and 3R-INN. In the bottom row sample, all methods struggle with accurate grain parameter prediction. Although FGA-CONVENT and FGA-NN introduce grain that does not match the original input, they enhance the overall visual quality of the reproduced video still. Note that in such cases, the interactive visualization tool presented in Section \ref{subsec:tool} could use the estimated parameters from FGA-NN and FGA-CONVENT as a starting point for further manual refinement, a capability absent in Style-FG and 3R-INN due to their latent space representations. The high computational cost of 3R-INN and Style-FG workflows prevents end-user device implementation, further highlighting the advantages of our learning-based analysis module coupled with a hardware-friendly synthesis module. Finally, encoding UHD videos with film grain at medium to low bitrates using our film grain analysis and synthesis workflow enables bitrate savings of up to 90\% compared to high-bitrate encoding \cite{jvet2}.

\section{Conclusion}
\label{sec:conclusion}
In this paper, we presented FGA-NN, the first learning-based film grain analysis method that provides normative film grain parameters from grainy videos, making it compatible with conventional film grain workflow in video standards. FGA-NN offers the best balance of prediction accuracy, compatibility, synthesis complexity, and generalization. This paper also introduces the first dataset of grainy videos paired with FGC-SEI parameters, ideal for training learning-based analysis methods. Future work will explore extending FGA-NN to other film grain parameters formats and codecs by adapting the training dataset. Subjective tests will also be conducted to further evaluate film grain fidelity.

\bibliographystyle{IEEEbib}
\bibliography{strings,refs}

\begin{thebibliography}{10}

\bibitem{vvc}
Benjamin Bross, Ye-Kui Wang, Yan Ye, Shan Liu, Jianle Chen, and Gary~J.
  Sullivan,
\newblock ``Overview of the versatile video coding (vvc) standard and its
  applications,''
\newblock {\em IEEE Transactions on Circuits and Systems for Video Technology},
  vol. 31, pp. 3736--3764, 2021.

\bibitem{sei}
Cristina Gomila and Alexander Kobilansky,
\newblock ``Sei message for film grain encoding,''
\newblock {\em ISO/IEC JTC1/SC29/WG11, ITU-T SG16 Q.6 document JVT-H022}, May
  2003.

\bibitem{milos}
Milo{\v{s}} Radosavljevic, Edouard Fran{\c{c}}ois, Erik Reinhard, Wassim
  Hamidouche, and Thomas Amestoy,
\newblock ``Implementation of film-grain technology within vvc,''
\newblock in {\em Applications of Digital Image Processing XLIV}. SPIE, 2021,
  vol. 11842, pp. 85--95.

\bibitem{stylefg}
Zoubida Ameur, Claire-H{\'e}l{\`e}ne Demarty, Olivier Le~Meur, Daniel
  M{\'e}nard, and Edouard Fran{\c{c}}ois,
\newblock ``Style-based film grain analysis and synthesis,''
\newblock in {\em Proceedings of the 14th Conference on ACM Multimedia
  Systems}, 2023, pp. 229--238.

\bibitem{3rinn}
Zoubida Ameur, Claire-H{\'e}l{\`e}ne Demarty, Daniel M{\'e}nard, and Olivier~Le
  Meur,
\newblock ``3r-inn: How to be climate friendly while consuming/delivering
  videos?,''
\newblock in {\em European Conference on Computer Vision}. Springer, 2025, pp.
  146--163.

\bibitem{vfgs}
{VFGS - Versatile film grain synthesis},
\newblock ``{https://github.com/InterDigitalInc/VersatileFilmGrain},''
\newblock {Accessed: 1/8/2024}.

\bibitem{VTMsoft}
{Versatile video coding (VVC) standard reference software VTM},
\newblock ``{https://vcgit.hhi.fraunhofer.de/jvet/VVCSoftware\_VTM},''
\newblock {Accessed: 18/1/2025}.

\bibitem{VVenC}
Adam Wieckowski, Jens Brandenburg, Tobias Hinz, Christian Bartnik, Valeri
  George, Gabriel Hege, Christian Helmrich, Anastasia Henkel, Christian
  Lehmann, Christian Stoffers, Ivan Zupancic, Benjamin Bross, and Detlev Marpe,
\newblock ``{VVenC: An Open And Optimized VVC Encoder Implementation},''
\newblock in {\em Proc. IEEE International Conference on Multimedia Expo
  Workshops (ICMEW)}, pp. 1--2.

\bibitem{VVdeC}
Adam Wieckowski, Gabriel Hege, Christian Bartnik, Christian Lehmann, Christian
  Stoffers, Benjamin Bross, and Detlev Marpe,
\newblock ``{Towards A Live Software Decoder Implementation For The Upcoming
  Versatile Video Coding (VVC) Codec},''
\newblock in {\em Proc. IEEE International Conference on Image Processing
  (ICIP)}, pp. 3124--3128.

\bibitem{bvi}
Di~Ma, Fan Zhang, and David~R Bull,
\newblock ``Bvi-dvc: A training database for deep video compression,''
\newblock {\em IEEE Transactions on Multimedia}, vol. 24, pp. 3847--3858, 2021.

\bibitem{div2k}
Eirikur Agustsson and Radu Timofte,
\newblock ``Ntire 2017 challenge on single image super-resolution: Dataset and
  study,''
\newblock in {\em The IEEE Conference on Computer Vision and Pattern
  Recognition (CVPR) Workshops}, July 2017.

\bibitem{Adam1}
Diederik~P Kingma and Jimmy Ba,
\newblock ``Adam: A method for stochastic optimization,''
\newblock {\em arXiv preprint arXiv:1412.6980}, 2014.

\bibitem{Adam2}
Sashank~J Reddi, Satyen Kale, and Sanjiv Kumar,
\newblock ``On the convergence of adam and beyond,''
\newblock {\em arXiv preprint arXiv:1904.09237}, 2019.

\bibitem{he2015delving}
Kaiming He, Xiangyu Zhang, Shaoqing Ren, and Jian Sun,
\newblock ``Delving deep into rectifiers: Surpassing human-level performance on
  imagenet classification,''
\newblock in {\em Proceedings of the IEEE international conference on computer
  vision}, 2015, pp. 1026--1034.

\bibitem{vtm}
{Rec. ITU-T H.266.2 | ISO/IEC 23090-16},
\newblock ``Reference software for itu-t h.266 versatile video coding,'' 2023,
\newblock Rec. ITU-T H.266.2 | ISO/IEC 23090-16.

\bibitem{jvet}
{Philippe de Lagrange},
\newblock ``Ahg4: Preparation of test material for film grain visual testing,''
  July 2024,
\newblock 35th Meeting, Sapporo, JP, 12–19 July 2024.

\bibitem{lpips}
Richard Zhang, Phillip Isola, Alexei~A Efros, Eli Shechtman, and Oliver Wang,
\newblock ``The unreasonable effectiveness of deep features as a perceptual
  metric,''
\newblock in {\em Proceedings of the IEEE conference on computer vision and
  pattern recognition}, 2018, pp. 586--595.

\bibitem{jsdnss}
Li-Heng Chen, Christos~G Bampis, Zhi Li, and Alan~C Bovik,
\newblock ``Learning to distort images using generative adversarial networks,''
\newblock {\em IEEE Signal Processing Letters}, vol. 27, pp. 2144--2148, 2020.

\end{thebibliography}

\end{document}